\documentclass{article}

\usepackage{microtype}
\usepackage{graphicx}
\usepackage{booktabs} 
\usepackage{amsfonts}
\usepackage{amsmath}
\usepackage{amsthm}
\usepackage{textcomp}

\usepackage{multirow}
\usepackage{enumitem}
\usepackage{bm}

\usepackage{graphicx, array, url}
\usepackage{tabularx,ragged2e}
\newcolumntype{C}{>{\Centering\arraybackslash}X} 

\usepackage{hyperref}

\usepackage[accepted]{icml2019}

\icmltitlerunning{Multiple instance learning with graph neural networks}

\begin{document}

\twocolumn[
\icmltitle{Multiple instance learning with graph neural networks}





\begin{icmlauthorlist}
\icmlauthor{Ming Tu}{to}
\icmlauthor{Jing Huang}{to}
\icmlauthor{Xiaodong He}{to}
\icmlauthor{Bowen Zhou}{to}
\end{icmlauthorlist}

\icmlaffiliation{to}{JD AI Research, Mountain View, CA 94043}

\icmlcorrespondingauthor{Ming Tu}{ming.tu@jd.com}

\icmlkeywords{Multiple instance learning, graph neural network, weakly-labeled supervised learning, differentiable pooling}

\vskip 0.3in
]



\printAffiliationsAndNotice{}  

\begin{abstract}
Multiple instance learning (MIL) aims to learn the mapping between a bag of instances and the bag-level label. In this paper, we propose a new end-to-end graph neural network (GNN) based algorithm for MIL: we treat each bag as a graph and use GNN to learn the bag embedding, in order to explore the useful structural information among instances in bags. The final graph representation is fed into a classifier for label prediction. 
Our algorithm is the first attempt to use GNN for MIL. We empirically show that the proposed algorithm achieves the state of the art performance on several popular MIL data sets without losing model interpretability.
\end{abstract}

\section{Introduction}
\label{intro}

Multiple instance learning (MIL) as a weakly-supervised learning algorithm deals with weakly-labeled data, where each data sample (often named as a bag) has multiple instances but only one label. MIL algorithms can be briefly categorized into three groups: instance-space algorithms \cite{ramon2000multi, raykar2008bayesian}, which compute a score for each instance as in single instance learning cases and then aggregate these scores for loss computation; bag-space algorithms, which directly calculate the similarity/distance between bags, and then employ lazy or kernel learning schemes to train the classifier \cite{wang2000solving, zhou2009multi, cheplygina2016dissimilarity, tu2017objective}; embedding-space algorithms, which convert the whole bag into a fixed-dimensional vector and then apply traditional single instance learning classifiers \cite{chen2006miles, wang2018revisiting, ilse2018attention}. It has been shown that the last two categories perform better than those on instance space in terms of bag-level accuracy, at the cost of losing the ability to detect key instances for model interpretation \cite{kandemir2015computer, ilse2018attention}. Recently deep neural networks (DNN) has been applied to MIL \cite{wang2018revisiting, ilse2018attention}. MIL algorithms based on DNN have achieved great improvement over the state of the art shallow learning algorithms. The basic idea is to do pooling operation on instance embeddings learned by DNN. Instead of untrainable pooling in \cite{wang2018revisiting}, attention mechanism was introduced in \cite{ilse2018attention} for pooling over instances, and the trainable attention weights on instances can provide extra information about the contribution of each instance to the final decision. Thus this approach is able to generate interpretable predictions.

However, most existing MIL algorithms treat instances in each bag as independently and identically distributed (i.i.d) samples \cite{zhou2007relation, zhou2009multi}. This strategy ignores the structural information presented among the instances in each bag. This assumption is not tenable in many situations. The experimental results in \cite{zhou2009multi} have shown that by constructing a graph for each bag and doing kernel learning on graphs is superior to those algorithms with i.i.d instance assumption. Furthermore, for tasks with sequential data, like document-level text classification where each document is a bag and sentences are instances, it is also unnatural to consider model input as uncorrelated instances \cite{angelidis2018multiple}.

Graph neural network (GNN) recently has attracted a lot of attention for learning tasks on structural data, for example node classification, link prediction and graph classification \cite{xu2018powerful}. The advantage of GNN is that it is able to efficiently and flexibly aggregate information through graph edges, and generate powerful representation of graph. In this paper, we propose a different paradigm to exploit the structural information in MIL. We assume that instances within a bag are correlated, and should not be treated as i.i.d samples. We regard each bag in MIL as a graph, and propose strategies to convert a bag of instances to an undirected graph. We then apply GNN for learning representation of bags in MIL. We make two major contributions: 1) Instead of graph kernel learning in \cite{zhou2009multi}, we apply GNN to learn the bag embedding. We show that by considering the structural information among multiple instances within bags, better bag representation can be achieved. Our proposed GNN-based MIL algorithm outperforms the state-of-the-art approaches measured in terms of classification accuracy on several popular MIL data sets. \footnote{Our code will be published after review.} 2) We further show that the proposed algorithm can also provide information about instances that are decisive to the final classification output. This retains model interpretability, which is important for health care applications.

\section{Methodology}

\subsection{How to apply GNN to MIL}

MIL can be formulated as a supervised learning task with bags of instances as input and bag-level labels as target. Given feature vectors of instances $X_i = [\mathbf{x}_1^{(i)},\mathbf{x}_2^{(i)},\cdots, \mathbf{x}_K^{(i)}]$ with all bags $[X_1, X_2, \cdots, X_N]$, the goal of MIL is to learn a mapping from all bags and their corresponding labels $[Y_1, Y_2,\cdots,Y_N]$. $N$ is the total number of bags, and $K$ is the number of instances in $i$-th bag. Note that $K$ can variate for different bags. The basic assumption of MIL is that if one bag at least has 1 positive example, then it is a positive bag; otherwise, it is a negative bag. While most of previous bag-space or embedding-space MIL algorithms assume that the instances within bags can be regarded as i.i.d samples, the studies in \cite{zhou2007relation, zhou2009multi} show that better performance can be achieved for both classification and regression tasks by considering the relational information among bag instances. This indicates that better bag representation can be derived by exploiting the structure information within bags in MIL.

GNN has shown great capability to do representation learning on graphs for either graph classification or node classification tasks. To make use of the relation information within bags in MIL, it is a good idea to treat each bag in MIL as a graph as what have been done in \cite{zhou2009multi}. To go further, we observe that graph classification and multiple instance learning are similar tasks if each bag in MIL is built into a graph: both of them take graph as input and output a graph label. To make this clear, we give a formal definition of graph based MIL:

\textit{
\textbf{Graph based MIL}: Given a set of bags $[X_1, X_2, \cdots, X_N]$, each of which contains multiple instances $[\mathbf{x}_1^{(i)},\mathbf{x}_2^{(i)},\cdots, \mathbf{x}_K^{(i)}]$
and a corresponding label $Y_i$, the goal is to learn the mappings: $\mathbb{X} \rightarrow \mathbb{G} \rightarrow \mathcal{Y}$, where $\mathbb{X}$ is the bag space, $\mathbb{G}$ is the graph space and $\mathcal{Y}$ is the label space. Each sample on graph space $\mathbb{G}$ is represented as a tuple $(A, V)$. $A \in \{0,1\}^{K \times K}$ is the adjacency matrix, and $V \in \mathbb{R}^{K \times D}$ is the feature matrix of all nodes. $D$ is the dimension of node feature.
}

While the mapping from bag space to graph space can be done heuristically (will be introduced in next subsection), the key of graph based MIL is how to learn the mapping from graph space to label space. Graph-level classification usually involves deriving a good representation of graphs given variant number of nodes and different graph structures, which requires to reduce the input graph to a fixed-dimensional feature vector. In this paper, we focus on GNN based graph representation learning for MIL, and propose a new angle to solve the MIL problem in the current study.

\subsection{Proposed algorithm}

\begin{figure}
    \centering
    \includegraphics[width=\linewidth]{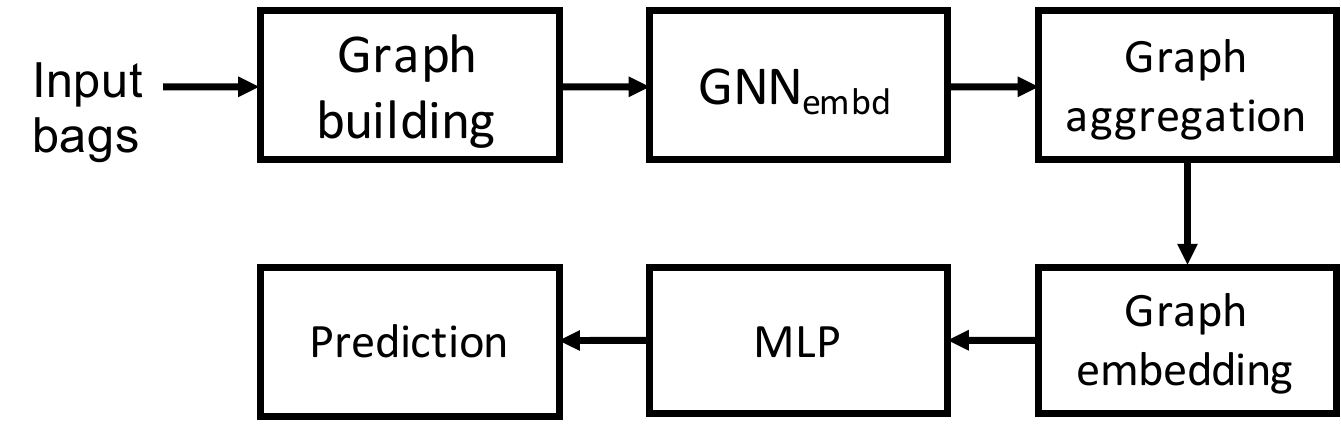}
    \caption{GNN based MIL framework overview.}
    \label{fig:my_label}
\end{figure}

Figure \ref{fig:my_label} illustrate the diagram of our proposed framework on GNN based MIL. First, to convert input bags of instances to graphs, we adopt a heuristic strategy similar with the one used in \cite{zhou2009multi}. Given a bag with instances $[\mathbf{x}_1^{(i)},\mathbf{x}_2^{(i)},\cdots, \mathbf{x}_K^{(i)}]$, the adjacency matrix $A$ can be derived with the following formula:

\begin{equation}
    A_{mn} =
    \begin{cases}
      1 & \text{if $dist(\mathbf{x}_m^{(i)}, \mathbf{x}_n^{(i)}) < \eta$} \\
      0 & \text{otherwise}
    \end{cases}
    \label{eq1}
\end{equation}
where $dist(\mathbf{x}_m^{(i)}, \mathbf{x}_n^{(i)})$ is the distance between $m$-th and $n$-th instance in bag $i$. In this study, Euclidean distance is employed for simplicity. $\eta$ is the threshold to decide whether there is an edge between two instances based on their distance. $\eta = 0$ means there is no edge in the input graph while $\eta = +\infty$ means the input is a complete graph. $\eta$ can be tuned for specific tasks.

After converting bags of instances to graphs, we propose an end-to-end graph representation learning algorithm based on GNN for MIL. Given an input graph $G_i$ with adjacency matrix $A_i \in \{0,1\}^{K \times K}$ and node feature matrix $V_i \in \mathbb{R}^{K \times D}$ constructed from a bag of $X_i$, a GNN is first applied to the input graph to conduct information passing over the graph. The output graph has the same number of nodes as the input graph, and the computation can be formulated as
\begin{equation}
    Z_i = GNN_{embd}(A_i, V_i),
\end{equation}
where $Z_i \in \mathbb{R}^{K \times D'}$ is the node embedding of graph output. $D'$ is the dimension of output node embedding, and can be different from input feature dimension $D$.

In order to obtain a fixed-dimensional representation of the graph, we need a strategy to aggregate information over the whole graph with adjacency matrix $A_i$ and updated node feature matrix $Z_i$. Inspired by its success on graph classification, the GNN based differentiable pooling algorithm \cite{ying2018hierarchical} is employed to collapse a graph with variant number of nodes to a vector representation. Differentiable pooling is composed of two operations: 1) learning an assignment matrix for the graph which gives the probability of a node belongs to a cluster. 2) collapsing graph nodes to the number of clusters by soft pooling given the learned assignment matrix. The number of clusters is predefined and the same for different graphs. The advantage of differentiable pooling is that it is able to learn the graph representation in a hierarchical way by doing graph clustering in multiple steps. 

Besides differentiable pooling based algorithm, we also implement an attention-based graph aggregation algorithm on top of $Z_i$, which is similar to the attention based MIL in \cite{ilse2018attention}, to show that our proposed paradigm is not limited to one specific graph representation learning algorithm. We will show the implementation details of both graph aggregation algorithms in supplementary materials.

\begin{table}[t]
\centering
\caption{Results on the five benchmark data sets. Bold numbers mean the highest average accuracy for that data set.}
\resizebox{1.0\columnwidth}{!}{%
\begin{tabular}{|c|c|c|c|c|c|}
\hline
Algorithms & MUSK1 & MUSK2 & FOX & TIGER & ELEPHANT \\ \hline
mi-Graph & 0.889\textpm0.033 & \textbf{0.903\textpm0.039} & 0.620\textpm0.044 & 0.860\textpm0.037 & 0.869\textpm0.035 \\ \hline
MI-Net & 0.887\textpm0.041 & 0.859\textpm0.046 & 0.622\textpm0.038 & 0.830\textpm0.032 & 0.862\textpm0.034 \\ \hline
MI-Net with DS & 0.894\textpm0.042 & 0.874\textpm0.043 & 0.630\textpm0.037 & 0.845\textpm0.039 & 0.872\textpm0.032 \\ \hline
Attention-MIL & 0.892\textpm0.040 & 0.858\textpm0.048 & 0.615\textpm0.043 & 0.839\textpm0.022 & 0.868\textpm0.022 \\ \hline
Attention-MIL with gating & 0.900\textpm0.050 & 0.863\textpm0.042 & 0.603\textpm0.029 & 0.845\textpm0.018 & 0.857\textpm0.027 \\ \hline
Ours & \textbf{0.917\textpm0.048} & 0.892\textpm0.011 & \textbf{0.679\textpm0.007} & \textbf{0.876\textpm0.015} & \textbf{0.903\textpm0.010} \\ \hline
\end{tabular}}
\label{table1}
\end{table}

\section{Experiments}
This section introduces the data sets and the performance of our proposed GNN based MIL algorithms. We compare our results with existing MIL algorithms. We also show the performance comparison between two implementations of GNN based MIL on a large medical image data set. The interpretability of our proposed model will be introduced in the last subsection.

\subsection{Five benchmark data sets}

In the first experiment, five most commonly used MIL data sets, which have been employed in almost all MIL studies, are adopted to show the proposed GNN based MIL algorithm can beat or compete with both DNN based MIL algorithms and traditional non-DNN MIL algorithms. Please refer to supplementary materials for details of the data sets. For fair comparison, we follow the same 10-fold cross validation (CV) as previous studies. Each data set is divided into 10 folds, and every time we use 9 folds for training and 1 fold for testing. We ran 5 times 10-fold CV with different random seeds. Our model is the differentiable pooling based algorithm as we found it works better than the attention based implementation. We calculate both the average and standard deviation of accuracy numbers, and compare them with previously proposed algorithms including ``mi-Graph''\cite{zhou2009multi}, ``MI-Net'' and ``Mi-Net with DS''\cite{wang2018revisiting}, ``Attention-MIL'' and ``Attention-MIL with gating''\cite{ilse2018attention} in Table \ref{table1}. The ``mi-Graph'' is based on kernel learning on graphs converted from bag of instances. The latter two algorithms are based on DNN and use either pooling or attention mechanism to derive the bag embedding. It can be seen from the results that the proposed GNN based MIL can give better results than previously algorithms on four data sets.

\begin{table}[t]
\centering
\caption{Results on 20 text categorization tasks. All results are averages of ten times running.}
\resizebox{1.0\columnwidth}{!}{%
\begin{tabular}{|c|c|c|c|c|}
\hline
tasks & mi-Graph & MI-Net & MI-Net with DS & Ours \\ \hline
alt.atheism & 0.655 & 0.776 & 0.860 & \textbf{0.863} \\ \hline
comp.graphics & 0.778 & 0.826 & 0.822 & \textbf{0.826} \\ \hline
comp.windows.misc & 0.631 & 0.678 & 0.716 & \textbf{0.726} \\ \hline
comp.ibm.pc.hardware & 0.595 & 0.778 & 0.792 & \textbf{0.794} \\ \hline
comp.sys.mac.hardware & 0.617 & 0.792 & 0.794 & \textbf{0.818} \\ \hline
comp.window.x & 0.698 & 0.786 & 0.812 & \textbf{0.828} \\ \hline
misc.forsale & 0.552 & 0.652 & 0.686 & \textbf{0.709} \\ \hline
rec.autos & 0.720 & 0.774 & 0.776 & \textbf{0.794} \\ \hline
rec.motorcycles & 0.640 & 0.762 & \textbf{0.868} & 0.838 \\ \hline
rec.sport.baseball & 0.647 & 0.856 & \textbf{0.874} & 0.844 \\ \hline
rec.sport.hockey & 0.850 & 0.862 & \textbf{0.912} & 0.883 \\ \hline
sci.crypt & 0.696 & 0.694 & \textbf{0.812} & 0.811 \\ \hline
sci.electronics & 0.871 & \textbf{0.930} & 0.926 & 0.918 \\ \hline
sci.med & 0.621 & 0.818 & \textbf{0.848} & 0.835 \\ \hline
sci.space & 0.757 & 0.752 & 0.818 & \textbf{0.860} \\ \hline
soc.religion.christian & 0.590 & 0.782 & \textbf{0.820} & 0.794 \\ \hline
talk.politics.guns & 0.585 & 0.652 & \textbf{0.780} & 0.773 \\ \hline
talk.politics.mideast & 0.736 & 0.794 & \textbf{0.842} & 0.840 \\ \hline
talk.politics.misc & 0.704 & 0.654 & 0.776 & \textbf{0.787} \\ \hline
talk.religion.misc & 0.633 & 0.700 & 0.758 & \textbf{0.782} \\ \hline
AVG & 0.679 & 0.766 & 0.815 & \textbf{0.816} \\ \hline
\end{tabular}}
\label{table2}
\end{table}

\subsection{Text categorization data sets}

In the second experiment, 20 text categorization data sets organized by authors of \cite{zhou2009multi} is used to verify the performance of GNN based MIL. Please refer to supplementary materials for details of the data sets. Since 10 different partitions of the 10 folds are already provided with the data set, we directly follow the same experimental design as in \cite{zhou2009multi, wang2018revisiting}. Our model is the differentiable pooling based algorithm as we found it works better than the attention based implementation.  In Table \ref{table2}, the results of ``mi-Graph'' and ``MI-Net'' based algorithms are compared with the proposed algorithm. The results show that ``MI-Net'' can achieve huge improvement over ``mi-Graph''. With DS, ``MI-Net'' can get further improvement. Our proposed GNN based MIL can beat the ``MI-Net with DS'' on 11 tasks, and the average accuracy on all 20 tasks is marginally better than the best performer on these data sets. The limited performance improvement is possibly due to that the instances within each bag is randomly selected.

\subsection{Retinal image classification}

A public available diabetic retinopathy screening data set called ``Messidor'' \cite{decenciere2014feedback} is adopted in the third experiment. Detecting diabetes from retinal image has attracted a lot of attention recently \cite{gulshan2016development}, and progress in this area is believed to have practical significance. The classification task using this data set is first formulated as a MIL problem in \cite{kandemir2015computer}. Please refer to supplementary materials for details of the data sets.

\begin{table}[]
\centering
\caption{Performance comparison between the proposed GNN based MIL algorithm and other existing MIL algorithms.}
\begin{tabular}{|c|c|c|}
\hline
Algorithms & Accuracy & F1 score \\ \hline
Ours-DP & \textbf{74.2}\% & \textbf{0.77} \\ \hline
Ours-Att & \textbf{72.9}\% & 0.75 \\ \hline
mi-Graph & 72.5\% & 0.75 \\ \hline
MILBoost & 64.1\% & 0.66 \\ \hline
Citation k-NN & 62.8\% & 0.68 \\ \hline
EMDD & 55.1\% & 0.69 \\ \hline
MI-SVM & 54.5\% & 0.70 \\ \hline
mi-SVM & 54.5\% & 0.71 \\ \hline
\end{tabular}
\label{table3}
\end{table}

\begin{table}[]
\caption{Comparison of TN (True Negative), FP (False Positive), FN (False Negative) and TP (True Positive) between models without graph input and with graph input.}
\resizebox{1.0\columnwidth}{!}{%
\begin{tabular}{|c|c|c|c|c|c|}
\hline
Models & TN & FP & FN & TP & Acc(\%) \\ \hline
Graph input & 379 & 167 & 143 & 511 & 74.2 \\ \hline
Without graph input & 374 & 172 & 159 & 495 & 72.4 \\ \hline
\end{tabular}}
\label{table4}
\end{table}

Two-fold CV is adopted to measure the proposed GNN based MIL as in \cite{kandemir2015computer}. We report both the accuracy and F1 score for this data set, and compare the results with those algorithms reported in \cite{kandemir2015computer}. In Table \ref{table3}, we compare the accuracy and F1 score of the proposed GNN-based algorithm with other algorithms, the numbers of performance measurements of which are from \cite{kandemir2015computer}. Except for ours, all other algorithms are non-DNN based algorithms and ``mi-Graph'' \cite{zhou2009multi} gave the best performance among them. We show that our proposed algorithm can further improve over the SOTA performance on this dataset with over 6\% relative reduction of the error rate and 2\% absolute improvement of the F1 score. We also show the performance of attention based implementation on this data set, which is worse than the differentiable pooling based implementation but still marginally better than existing algorithms.

In Table \ref{table4}, we compare the results between model with graph input and model without graph input ($\eta$=0 in equation \ref{eq1}). It shows obvious improvement over the model without graph input and the performance elevation mainly comes from less false negatives. This further proves that the model being able to exploit structural information among instances can achieve better performance.

\subsection{Model interpretability}

\begin{figure}
    \includegraphics[width=\linewidth]{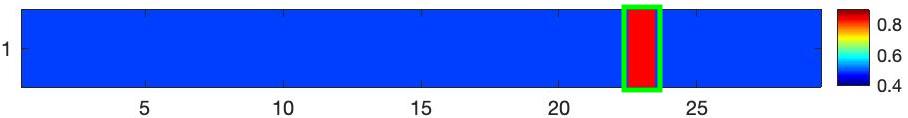} \\
    \includegraphics[width=\linewidth]{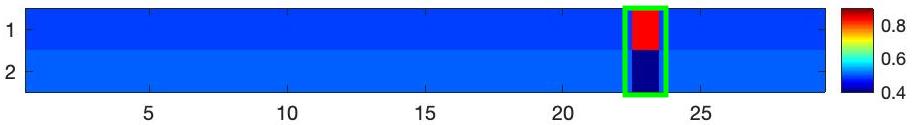} \\
    \includegraphics[width=\linewidth]{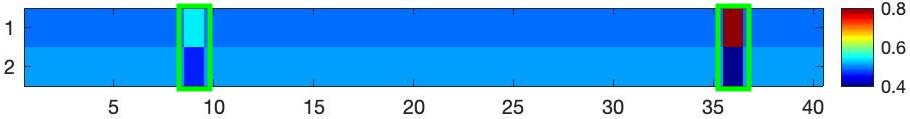}
    \caption{Heat maps of learned assignment matrices of different bags. X axis indicates the indices of instances and Y axis indicates the cluster indices. Instances within green box are positive instances in that bag. The values of colormap represent the probability that an instance belongs to a cluster. In order to make the figure more contrasting, we adjust the range of colormap values.}
    \label{heatmap}
\end{figure}

In Figure \ref{heatmap}, we show some heat maps of the learned assignment matrices of bags returned by the differentiable pooling algorithm on text categorization tasks (details about calculation of assignment matrices will be provided in supplementary materials). We choose text categorization tasks because the ground-truth instance labels are provided. In Figure \ref{heatmap}, we show three heat maps of the learned assignment matrices of different bags. These heat maps show that our model is able to locate important instances or separate positive and negative instances in MIL bags. This analysis proves that our proposed GNN based MIL retains the model interpretability in contrast to the study in \cite{zhou2009multi}.

\section{Conclusion}
In this paper, a new GNN based paradigm is proposed for tackling MIL problems. Instead of regarding instances in MIL bags as i.i.d samples, we first convert each bag of instances into a graph, and then use an end-to-end GNN based network to learn the representation of the graph as the embedding of the bag. The benefit of treating each bag as instances is that relation information among instances can be exploited for specific learning tasks. Finally, the learned graph representation is fed into a MLP based classifier to predict the bag-level labels. Through experiments on different sets of data, we show that the proposed GNN based MIL is able to achieve the state-of-the-art performance on the popular MIL data sets, thus proves its superiority over existing methods while retaining model interpretability.




\nocite{langley00}

\bibliography{example_paper}
\bibliographystyle{icml2019}

\twocolumn[
\icmltitle{Supplementary materials}



\icmlsetsymbol{equal}{*}





\vskip 0.3in
]




\section{Algorithm details}
\label{intro}

\subsection{Differentiable pooling based algorithm}

Another GNN is applied on top of $Z_i$ in main text but with different purpose:
\begin{equation}
    S_i = softmax(GNN_{cluster}(A_i, V_i)),
    \label{eq6}
\end{equation}
where $GNN_{cluster}$ functions like a dimension reduction module and the output dimension of $GNN_{cluster}$ is $K \times C$, where $C$ is the predefined number of clusters. The $softmax$ function converts the output to probabilities.

The soft pooling step takes node embedding $Z_i$ and node assignment matrix $S_i$ as input. Then it generates a coarsened graph with $C$ nodes by re-calculating the node embeddings and adjacency matrix as follows:
\begin{equation}
    V_i^* = S_i^T Z_i \in \mathbb{R}^{C \times D'},
    \label{eq7}
\end{equation}
\begin{equation}
    A_i^* = S_i^T A_i S_i \in \mathbb{R}^{C \times C}.
    \label{eq8}
\end{equation}
$V_i^*$ and $A_i^*$ defines the node feature matrix and adjacency matrix of the new graph with $C$ nodes. $C$ belongs to the hyperparameters of the model, and can be tuned based on tasks. For example, if $C$ is set to 1, then the coarsened graph has only one node, the embedding of which can be regarded as the learned graph representation. if $C$ is set to 2, then an extra operation such as max pooling or concatenation can be applied to get the graph embedding. Also, the differentiable pooling can be done for multiple steps in a hierarchical way as shown in \cite{ying2018hierarchical}. However, for MIL applications, the number of instances $K$ in each bag can be as small as 1. Thus, the number of steps for differential pooling and the number of clusters should be adjusted accordingly.

In this study, the GNN module is a variant of the GraphSAGE proposed in \cite{hamilton2017inductive}, which combines the ``aggregation'' and ``combination'' steps into one formula:
\begin{equation}
 \resizebox{0.88\hsize}{!}{
    $\bm{v}_k \leftarrow act(W \cdot MEAN(\bm{v}_u, \forall u \in \mathcal{N}(k) \cup \{k\}))$,
    }
\end{equation}
where $\mathcal{N}(k)$ is the neighbors of node $k$. $act$ is the activation function and is implemented with LeakyReLU (leaky rectified linear unit as activation function).

\begin{figure}[t]
    \centering
    \includegraphics[width=\linewidth]{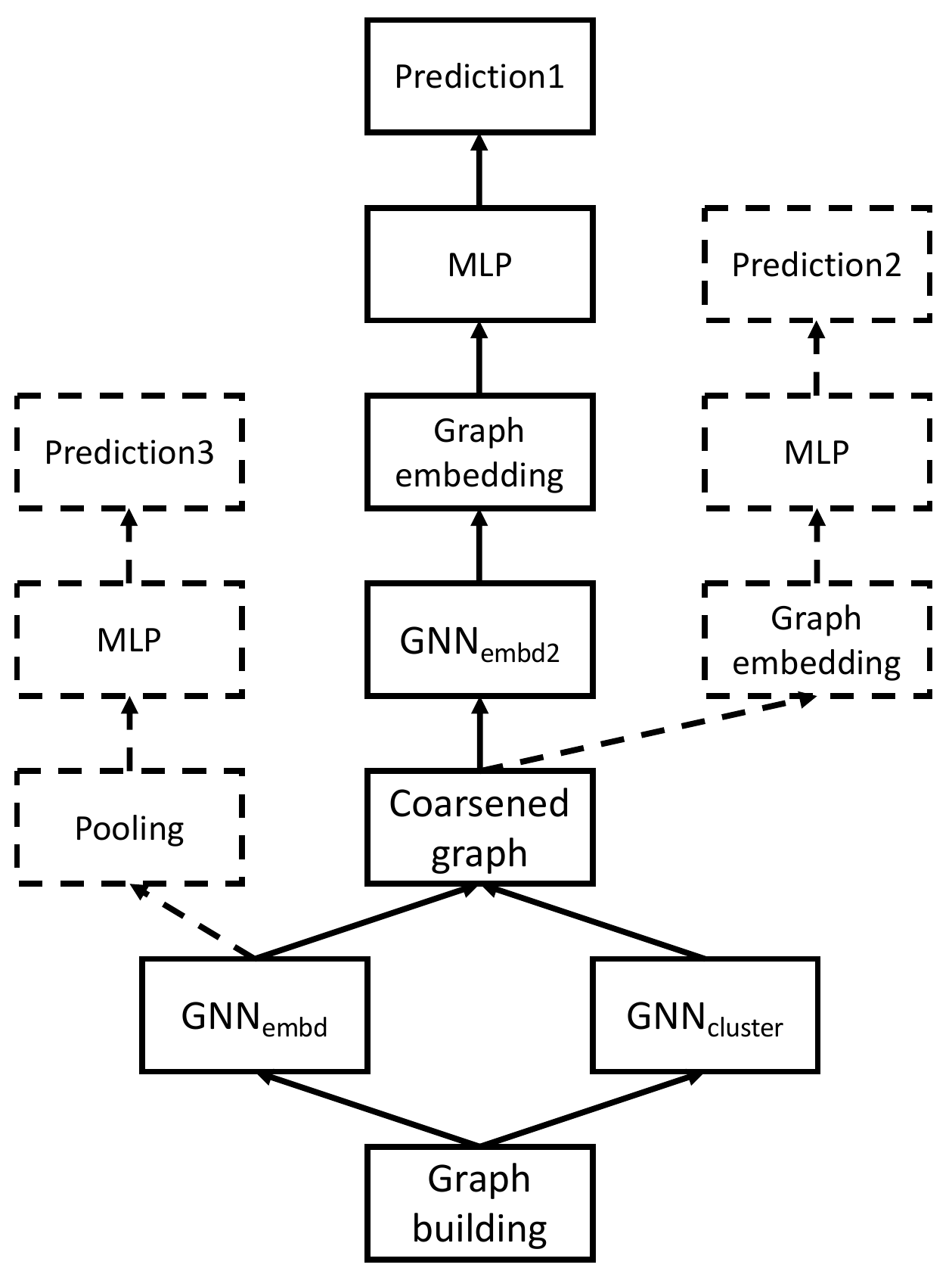}
    \caption{Network architecture for the proposed GNN based MIL}
    \label{network}
\end{figure}

Besides $GNN_{embd}$ and $GNN_{cluster}$, we apply another GNN (annotated as ``$GNN_{embd2}$'') to the output of differentiable pooling for an extra step of information passing on the small graph. Then either max pooling or concatenaion can be applied to the output of ``$GNN_{embd2}$'' depending on the predefined number of clusters $C$. After getting the fixed-dimensional graph embedding, it will be fed to a classifier based on Multiple Layer Perceptron (MLP). The number of output nodes depends on the number of classes of the classification task. Also, we find that the deep supervision (DS) technique proposed in \cite{lee2015deeply} is also beneficial to our proposed algorithm as in \cite{wang2018revisiting}. The motivation of DS is that it works as a kind of regularization and can relieve gradient vanishing problem thus making training more stable. Specifically, in addition to the loss calculated at the MLP output, we apply a pooling operation to the output of both $GNN_{embd}$ and differentiable pooling, and include the loss at those two places to the final training loss.

Figure \ref{network} illustrates the detailed network architecture of the proposed GNN based MIL algorithm. The input bags are first converted to graphs by the ``graph building'' module. The ``GNN\textsubscript{embd}'' is for updating the original input node features, and ``GNN\textsubscript{cluster}'' is for learning the cluster assignment matrix. Then, applying equation \ref{eq7} and \ref{eq8} generate a coarsened graph with number of nodes the same as the predefined number of clusters of ``GNN\textsubscript{cluster}''. ``GNN\textsubscript{embd2}'' further updates the node features of the coarsened graphs, then converts the node embeddings to graph embedding with different strategies (max-pooling or concatenation). MLP is employed to make the prediction on input bag given its graph embedding. The modules of blocks with dashed lines indicate the application of deep supervision (DS). We found that by adding DS the performance can be improved. So, the final cross entropy loss is computed on ``prediction1'', ``prediction2'' and ``prediction3'' (weight for each item can be tuned). More details about the hyperparameters of different modules, for example the number of hops of different GNN modules, will be introduced in supplementary materials.

\subsection{Attention based algorithm}

We also implement another strategy to get graph representation of $Z_i$. The idea is similar to the attention based MIL algorithm proposed in \cite{ilse2018attention}. The difference is that the attention based MIL still consider the instances in each bag as i.i.d samples, and use a self-attention mechanism to do weighted sum of instance features, which then yield a fixed-dimensional embedding for each bag. However, we apply attention to the output of GNN after message passing among nodes of the input graph ($Z_i$ in equation 2 in main text). Formally, assume $j$-th node feature $\mathbf{z}_i^j$ of $i$-th bag feature matrix $Z_i$, the graph embedding can be calculated with the following equations:

\begin{equation}
    V_i^* = \sum_{}{\mathbf{\alpha}_i^j \mathbf{z}_i^j},
\end{equation}
where $\mathbf{\alpha}_i^j$ is obtained by:
\begin{equation}
    \mathbf{\alpha}_i^j = softmax(MLP_{att}(\mathbf{z}_i^j)).
\end{equation}
$softmax()$ converts the output of $MLP_{att}()$ to a probability by normalizing over all instances.

With the attention based graph embedding $V_i^*$, we can add multiple layers of MLPs to get a prediction of the bag label. Similarly, we use the same DS technique as in the last subsection. It is reasonable to note that the differential pooling based graph aggregation utilize the graph relation information during the aggregation process while the attention based algorithm does not.

\section{Data sets and more details of experiments}

The proposed GNN based MIL is evaluated with different sets of data, which contain five popular MIL data sets including drug activity prediction and image classification, 20 text categorization tasks used in \cite{zhou2009multi,wang2018revisiting} and a medical image classification task for diagnosing diabetes from weakly labeled retinal images. We aim to show that: 1) the proposed algorithm is superior to both strong DNN based MIL baselines \cite{wang2018revisiting, ilse2018attention} that consider bag instances as i.i.d samples, and kernel learning based methods \cite{zhou2009multi} that is able to utilize the relation information among bag instances. 2) the proposed algorithm also retain the ability to interpret the model output by giving information examples that is decisive to the final prediction. Note that since the number of instances per bag in these data sets can be as small as 1, for differentiable pooling based algorithm we only do one step differentiable pooling and the number of clusters is chosen from \{1, 2\}.

\begin{table*}[t]
\centering
\caption{Network module configuration}
\setlength\extrarowheight{2pt} 
\begin{tabularx}{\textwidth}{|C|C|}
\hline
 Modules & Configurations \\ \hline
 GNN\textsubscript{embd} & 1 layer GraphSage followed by leaky ReLU activation function (with negative slope equals to 0.01) and batch normalization. Input and output feature dimensions are the same.  \\ \hline
 GNN\textsubscript{pool} & 1 layer GraphSage followed by leaky ReLU activation function (with negative slope equals to 0.01) and batch normalization. 1 extra layer of MLP with leaky ReLU activation function applied to the output of GraphSage. Input and output feature dimensions are the same.  \\ \hline
 GNN\textsubscript{embd2} & Same as GNN\textsubscript{embd} \\ \hline
 MLP & 2 layers of MLP with leaky ReLU activation function. The output dimension of 1 layer is the half of the input dimension. \\ \hline
\end{tabularx}
\label{table1}
\end{table*}

\subsection{Datasets}
For the five benchmark data sets, MUSK1 has 47 positive bags and 45 negative bags while MUSK2 has 39 positive bags and 63 positive bags. Instances within a bag are the different conformations of a molecule. The task is to predict whether new molecules will be musks or non-musks. All three image classification data sets have 100 positive examples and 100 negative examples. Each image has several regions of interest (ROI) which are regarded as instances. The goal is to predict whether new images belong to the target categorization.
The five benchmark data sets and the Messidor data set used in this study can be downloaded online \footnote{\url{https://figshare.com/articles/MIProblems\_A_repository\_of\_multiple\_instance\_learning\_datasets/6633983} (Credit to Veronika Cheplygina)}. 

For the 20 text categorization data sets, each one has 50 positive bags and 50 negative bags. There are about 3\% positive instances in the target category in positive bags while negative instances are drawn from other categories. Every instance is represented by the top 200 term frequency–inverse document frequency (TF-IDF) features. The task is to classify a given bag of texts into target categorization. The text categorization data sets can be downloaded online \footnote{\url{http://lamda.nju.edu.cn/data_MItext.ashx}}.

Messidor is a weakly labeled data set with 654 positive (diagnosed with diabetes) and 546 negative (healthy) images. The size of each image is 700 $\times$ 700 pixels. Each image is partitioned into small patches of 135 $\times$ 135 pixels. Patches with only background are dropped. There are 12352 instances in total, each of which is represented with a 687 dimensional feature vectors. Features contain intensity histogram of RGB channels for 26 bins, mean of local binary pattern histograms of 20 $\times$ 20 pixel grids, mean of SIFT descriptors, and box count for grid sizes 2 to 8. The task for this data set is to detect whether the given image is from healthy subject or subject with diabetes.

\subsection{Implementation details}

Our implementation is based on the open source geometric deep learning library Pytorch geometric\footnote{\url{https://github.com/rusty1s/pytorch_geometric}}, which is based on Pytorch. We will make our code open source after the reviewing process. In table \ref{table1}, we show the detailed configuration of our network modules ``GNN\textsubscript{embd}'', ``GNN\textsubscript{pool}'', ``GNN\textsubscript{embd2}'' and ``MLP'' for differentiable pooling based algorithm. We also use the same link prediction regularization as in the original differentiable pooling paper introduced in the main text. For the attention based algorithm, it is more straight forward so we do not include the details here. 

The hyperparameters include threshold $\eta$ when building graphs from input bags, batch size, number epochs, learning rate, weight decay, number of cluster for GNN\textsubscript{pool}, max pooling or concatenation when deriving graph embedding if number of cluster is larger than 1, weight for regularization (we use same weight for the three parts of loss in deep supervision). We also use a cosine annealing strategy for learning rate schedule. For examples, the hyperparameters of the Messidor experiments are set to: threshold $\eta$ (+$\infty$), batch size (128), number epochs (50), learning rate (3e-4), weight decay (1e-3), number of cluster (1), weight for regularization (0.5). All experiments run on a single GPU.

\subsection{Model interpretability}

The ability to identify instances that are decisive to the class prediction can make MIL algorithms more practical in real applications, especially for health care applications \cite{ilse2018attention}. In this subsection, we will show that the proposed GNN based MIL with differentiable pooling algorithm is able to identify decisive instances that contribute more to the final decision. The interpretability of attention based algorithm is already explored in \cite{ilse2018attention}. The most key part of differentiable pooling is the learned assignment matrix $S_i$ in equation \ref{eq6}, which gives the probability of each node belongs to a cluster. $S_i$ is further employed in equations \ref{eq7} and \ref{eq8} to cluster graph nodes into different clusters. It is reasonable to assume the following cluster assignment could happen: 1) if we choose to collapse the graph nodes into 2 clusters, then positive instances could be in the same cluster while negative instances could be in the other cluster. 2) if we choose to collapse all graph nodes into 1 cluster, then we expect that positive instances could be given higher probabilities than negative instances, which functions like the attention mechanism in \cite{ilse2018attention}. The reason for these assumptions is that the graphs are built upon pairwise euclidean distances among nodes, thus they probably will be close on the built graph and will possibly stay in the same cluster \cite{ying2018hierarchical}. If there is only one cluster, then the assignment matrix $S_i$ tends to make positive examples stand out because $S_i$ is learned to maximize the classification accuracy and positive instances are those push the classifier to learn the proper model parameters.





\bibliography{workshop_paper}

\begin{thebibliography}{18}
\providecommand{\natexlab}[1]{#1}
\providecommand{\url}[1]{\texttt{#1}}
\expandafter\ifx\csname urlstyle\endcsname\relax
  \providecommand{\doi}[1]{doi: #1}\else
  \providecommand{\doi}{doi: \begingroup \urlstyle{rm}\Url}\fi

\bibitem[Angelidis \& Lapata(2018)Angelidis and Lapata]{angelidis2018multiple}
Angelidis, S. and Lapata, M.
\newblock Multiple instance learning networks for fine-grained sentiment
  analysis.
\newblock \emph{Transactions of the Association of Computational Linguistics},
  6:\penalty0 17--31, 2018.

\bibitem[Chen et~al.(2006)Chen, Bi, and Wang]{chen2006miles}
Chen, Y., Bi, J., and Wang, J.~Z.
\newblock Miles: Multiple-instance learning via embedded instance selection.
\newblock \emph{IEEE Transactions on Pattern Analysis and Machine
  Intelligence}, 28\penalty0 (12):\penalty0 1931--1947, 2006.

\bibitem[Cheplygina et~al.(2016)Cheplygina, Tax, and
  Loog]{cheplygina2016dissimilarity}
Cheplygina, V., Tax, D.~M., and Loog, M.
\newblock Dissimilarity-based ensembles for multiple instance learning.
\newblock \emph{IEEE Trans. Neural Netw. Learning Syst.}, 27\penalty0
  (6):\penalty0 1379--1391, 2016.

\bibitem[Decenci{\`e}re et~al.(2014)Decenci{\`e}re, Zhang, Cazuguel, Lay,
  Cochener, Trone, Gain, Ordonez, Massin, Erginay,
  et~al.]{decenciere2014feedback}
Decenci{\`e}re, E., Zhang, X., Cazuguel, G., Lay, B., Cochener, B., Trone, C.,
  Gain, P., Ordonez, R., Massin, P., Erginay, A., et~al.
\newblock Feedback on a publicly distributed image database: the messidor
  database.
\newblock \emph{Image Analysis \& Stereology}, 33\penalty0 (3):\penalty0
  231--234, 2014.

\bibitem[Gulshan et~al.(2016)Gulshan, Peng, Coram, Stumpe, Wu, Narayanaswamy,
  Venugopalan, Widner, Madams, Cuadros, et~al.]{gulshan2016development}
Gulshan, V., Peng, L., Coram, M., Stumpe, M.~C., Wu, D., Narayanaswamy, A.,
  Venugopalan, S., Widner, K., Madams, T., Cuadros, J., et~al.
\newblock Development and validation of a deep learning algorithm for detection
  of diabetic retinopathy in retinal fundus photographs.
\newblock \emph{Jama}, 316\penalty0 (22):\penalty0 2402--2410, 2016.

\bibitem[Hamilton et~al.(2017)Hamilton, Ying, and
  Leskovec]{hamilton2017inductive}
Hamilton, W., Ying, Z., and Leskovec, J.
\newblock Inductive representation learning on large graphs.
\newblock In \emph{Advances in Neural Information Processing Systems}, pp.\
  1024--1034, 2017.

\bibitem[Ilse et~al.(2018)Ilse, Tomczak, and Welling]{ilse2018attention}
Ilse, M., Tomczak, J.~M., and Welling, M.
\newblock Attention-based deep multiple instance learning.
\newblock \emph{arXiv preprint arXiv:1802.04712}, 2018.

\bibitem[Kandemir \& Hamprecht(2015)Kandemir and
  Hamprecht]{kandemir2015computer}
Kandemir, M. and Hamprecht, F.~A.
\newblock Computer-aided diagnosis from weak supervision: a benchmarking study.
\newblock \emph{Computerized medical imaging and graphics}, 42:\penalty0
  44--50, 2015.

\bibitem[Lee et~al.(2015)Lee, Xie, Gallagher, Zhang, and Tu]{lee2015deeply}
Lee, C.-Y., Xie, S., Gallagher, P., Zhang, Z., and Tu, Z.
\newblock Deeply-supervised nets.
\newblock In \emph{Artificial Intelligence and Statistics}, pp.\  562--570,
  2015.

\bibitem[Ramon \& De~Raedt(2000)Ramon and De~Raedt]{ramon2000multi}
Ramon, J. and De~Raedt, L.
\newblock Multi instance neural networks.
\newblock In \emph{Proceedings of the ICML-2000 workshop on attribute-value and
  relational learning}, pp.\  53--60, 2000.

\bibitem[Raykar et~al.(2008)Raykar, Krishnapuram, Bi, Dundar, and
  Rao]{raykar2008bayesian}
Raykar, V.~C., Krishnapuram, B., Bi, J., Dundar, M., and Rao, R.~B.
\newblock Bayesian multiple instance learning: automatic feature selection and
  inductive transfer.
\newblock In \emph{Proceedings of the 25th international conference on Machine
  learning}, pp.\  808--815. ACM, 2008.

\bibitem[Tu et~al.(2017)Tu, Berisha, and Liss]{tu2017objective}
Tu, M., Berisha, V., and Liss, J.
\newblock Objective assessment of pathological speech using distribution
  regression.
\newblock In \emph{Acoustics, Speech and Signal Processing (ICASSP), 2017 IEEE
  International Conference on}, pp.\  5050--5054. IEEE, 2017.

\bibitem[Wang \& Zucker(2000)Wang and Zucker]{wang2000solving}
Wang, J. and Zucker, J.-D.
\newblock Solving multiple-instance problem: A lazy learning approach.
\newblock 2000.

\bibitem[Wang et~al.(2018)Wang, Yan, Tang, Bai, and Liu]{wang2018revisiting}
Wang, X., Yan, Y., Tang, P., Bai, X., and Liu, W.
\newblock Revisiting multiple instance neural networks.
\newblock \emph{Pattern Recognition}, 74:\penalty0 15--24, 2018.

\bibitem[Xu et~al.(2018)Xu, Hu, Leskovec, and Jegelka]{xu2018powerful}
Xu, K., Hu, W., Leskovec, J., and Jegelka, S.
\newblock How powerful are graph neural networks?
\newblock \emph{arXiv preprint arXiv:1810.00826}, 2018.

\bibitem[Ying et~al.(2018)Ying, You, Morris, Ren, Hamilton, and
  Leskovec]{ying2018hierarchical}
Ying, Z., You, J., Morris, C., Ren, X., Hamilton, W., and Leskovec, J.
\newblock Hierarchical graph representation learning with differentiable
  pooling.
\newblock In \emph{Advances in Neural Information Processing Systems}, pp.\
  4805--4815, 2018.

\bibitem[Zhou \& Xu(2007)Zhou and Xu]{zhou2007relation}
Zhou, Z.-H. and Xu, J.-M.
\newblock On the relation between multi-instance learning and semi-supervised
  learning.
\newblock In \emph{Proceedings of the 24th international conference on Machine
  learning}, pp.\  1167--1174. ACM, 2007.

\bibitem[Zhou et~al.(2009)Zhou, Sun, and Li]{zhou2009multi}
Zhou, Z.-H., Sun, Y.-Y., and Li, Y.-F.
\newblock Multi-instance learning by treating instances as non-iid samples.
\newblock In \emph{Proceedings of the 26th annual international conference on
  machine learning}, pp.\  1249--1256. ACM, 2009.

\end{thebibliography}
\bibliographystyle{icml2019}

\end{document}